\pdfoutput=1
\documentclass[11pt]{article}

\usepackage{acl}

\usepackage{times}
\usepackage{latexsym}
\usepackage{comment}
\usepackage{graphicx}
\usepackage{nicefrac}
\usepackage[ruled,vlined,linesnumbered,noend]{algorithm2e}
\usepackage{xcolor}

\usepackage[T1]{fontenc}
\usepackage[utf8]{inputenc}
\definecolor{darkgreen}{RGB}{0,160,0}

\usepackage{microtype}

\title{Towards Fair Evaluation of Dialogue State Tracking by Flexible Incorporation of Turn-level Performances}

\author{Suvodip Dey, Ramamohan Kummara,  Maunendra Sankar Desarkar\\
   Indian Institute of Technology Hyderabad, India \\
   \texttt{\small{cs19resch01003@iith.ac.in,  cs19mds11004@iith.ac.in,  maunendra@cse.iith.ac.in}} \\
   }

\begin{document}
\maketitle

\begin{abstract}
% Dialogue State Tracking (DST) is primarily evaluated using Joint Goal Accuracy (JGA) which is the fraction of turns where the ground-truth dialogue state exactly matches the prediction. Generally in DST, the dialogue state or belief state for a given turn contain all the intents shown by the user till that turn. Due to this cumulative nature of the belief state, it is difficult to get a correct prediction once a misprediction has occurred. Thus, although being a useful metric, it can be harsh at times and underestimate the true potential of a DST model. Moreover, it has been observed that improvement in JGA can sometimes decrease the performance of turn-level or non-cumulative belief state prediction. So, using JGA as the only metric for model selection may not be ideal for all scenarios. In this work, we discuss various evaluation metrics used for DST along with their shortcomings. To address the existing issues, we propose a new evaluation metric named \textbf{F}lexible \textbf{G}oal \textbf{A}ccuracy (\textbf{FGA}). FGA is a generalized version of JGA. But unlike JGA, it tries to give penalized credit to mispredictions that are locally correct but the root cause of the error is an earlier turn. By doing so, FGA considers both cumulative and turn-level belief states in a flexible manner and provides a better insight in comparison to the existing metrics. We report the performance of different DST models on the MultiWOZ 2.1 dataset.
Dialogue State Tracking (DST) is primarily evaluated using Joint Goal Accuracy (JGA) defined as the fraction of turns where the ground-truth dialogue state exactly matches the prediction. Generally in DST, the dialogue state or belief state for a given turn contains all the intents shown by the user till that turn. Due to this cumulative nature of the belief state, it is difficult to get a correct prediction once a misprediction has occurred. Thus, although being a useful metric, it can be harsh at times and underestimate the true potential of a DST model. Moreover, an improvement in JGA can sometimes decrease the performance of turn-level or non-cumulative belief state prediction due to inconsistency in annotations. So, using JGA as the only metric for model selection may not be ideal for all scenarios. In this work, we discuss various evaluation metrics used for DST along with their shortcomings. To address the existing issues, we propose a new evaluation metric named \textbf{F}lexible \textbf{G}oal \textbf{A}ccuracy (\textbf{FGA}). FGA is a generalized version of JGA. But unlike JGA, it tries to give penalized rewards to mispredictions that are locally correct i.e. the root cause of the error is an earlier turn. By doing so, FGA considers the performance of both cumulative and turn-level prediction flexibly and provides a better insight than the existing metrics. We also show that FGA is a better discriminator of DST model performance.
% We also show that FGA has a better correlation with human ratings than JGA.
\end{abstract}

\section{Introduction}
Dialogue State Tracking (DST) is at the core of task-oriented dialogue systems. It is responsible for keeping track of the key information exchanged during a conversation. With the growing popularity of task-based conversational agents, it is essential to review the evaluation of DST to appropriately measure the progress in this evolving area.

% The task of DST is to predict the user intent through dialogue states \cite{dstc2}. Fig \ref{fig:dst-pr} shows an example DST task from MultiWOZ \cite{multiwoz} dataset. Let $U_t$ and $S_t$ be the user and system utterances respectively at turn $t$. Then a typical conversation can be expressed as $D = \{ U_0, (S_1,U_1),...(S_{n},U_{n})\}$. The commonly used ground-truth dialogue state for DST is the belief state. Belief state $B_t$ for turn $t$ is defined as the set of \textit{(domain, slot, slot-value)} triplets that have been extracted till turn $t$. It is to be noted that the belief state for a given turn is cumulative. The objective of DST is to predict $B_t$ given the dialogue history till turn $t$.
The task of DST is to predict the user intent through dialogue states \cite{dstc2}. Fig. \ref{fig:dst-pr} shows an example DST task from MultiWOZ \cite{multiwoz} dataset. Let $U_t$ and $S_t$ be the user and system utterances respectively at turn $t$. Then a typical conversation can be expressed as $D = \{ U_0, (S_1,U_1),...(S_{n},U_{n})\}$. The commonly used ground-truth dialogue state for DST is the belief state. Belief state $B_t$ for turn $t$ is defined as the set of \textit{(domain, slot, slot-value)} triplets that have been extracted till turn $t$, thereby it is cumulative in nature. The objective of DST is to predict $B_t$ given the dialogue history till turn $t$.

% DST is broadly evaluated using joint goal accuracy (Section \ref{sec:joint_acc}), slot accuracy (Section \ref{sec:slot_acc}), and average goal accuracy (Section \ref{sec:avg_acc}). However, the primary metric for evaluating DST is joint goal accuracy (JGA) because it fits well with the use case. JGA compares the predicted dialogue states to the ground truth $B_t$ at each dialogue turn $t$. As the belief state is cumulative, it is very unlikely for a model to get back a correct prediction after a misprediction. This is why JGA is a strict metric and can provide an underestimated performance in certain cases. Besides, JGA only focuses on the performance of the cumulative belief state prediction and completely ignores the performance of turn-specific local prediction. Let $T_t$ be the turn-level belief state that contains all the intents or \textit{(domain, slot, slot-value)} triplets expressed by the user only at turn $t$. Ideally, a model with higher JGA should also perform well to predict $T_t$. Unfortunately, we observe that improving JGA can sometimes degrade the performance of predicting $T_t$. It is not an issue if the end goal is only to improve the prediction of $B_t$. But there can be scenarios where the objective is to simultaneously track the performance of both $B_t$ and $T_t$. In such cases, there is a requirement for a better alternative than JGA.
The primary metric for evaluating DST is Joint Goal Accuracy (JGA). It compares the predicted dialogue states to the ground truth $B_t$ at each dialogue turn $t$ \cite{dstc2}. As the belief state is cumulative, it is very unlikely for a model to get back a correct prediction after a misprediction. This is why it can provide an underestimated performance in certain cases. Besides, JGA completely ignores the performance of turn-specific local predictions. Let $T_t$ be the turn-level belief state that contains all the intents or \textit{(domain, slot, slot-value)} triplets expressed by the user only at turn $t$. Ideally, a model with higher JGA should also perform equally well to predict $T_t$. But, we observe that improving JGA can sometimes degrade the performance of predicting $T_t$ mainly due to the presence of annotation inconsistencies in the available datasets. For example, in Fig. \ref{fig:dst-pr}, the presence of \textit{(hotel, area, centre)} and absence of \textit{(attraction, name, all saints church)} in ground-truth $B_2$  and $B_4$ shows such inconsistencies. So, the generalization of the model may get compromised if the model selection is done only using JGA. Annotation inconsistencies and errors are common in real-world datasets. Hence, to provide a fair estimate, it requires not only track the performance of the cumulative belief state but also turn-level belief state as well.

In this work, we address these issues of JGA by proposing a novel evaluation metric for DST called \textbf{F}lexible \textbf{G}oal \textbf{A}ccuracy (\textbf{FGA}). The central idea of FGA is to partially penalize a misprediction which is locally correct i.e. the source of the misprediction is some earlier turn. The main contributions of our work are as follows
\footnote[1]{Code is available at \href{https://github.com/SuvodipDey/FGA}{github.com/SuvodipDey/FGA}}:
% \begin{itemize}
%     \item We provide a detailed analysis of the existing metrics to evaluate DST.
%     \item We introduce Flexible Goal Accuracy (FGA), a new metric to evaluate DST. Unlike the existing metrics, FGA can keep track of the performance of both cumulative and non-cumulative belief state prediction simultaneously.
%     % \item By construction of the metric, joint goal accuracy becomes a special case of Flexi. 
%     \item We perform a human evaluation of a DST model and show that FGA has a slightly better correlation with the human ratings in comparison to JGA.
% \end{itemize}
\begin{itemize}
    \item Detailed analysis of the existing DST metrics.
    \item Proposal of Flexible Goal Accuracy (FGA) than can keep track of both joint and turn-level performances simultaneously.
    \item Justification of FGA along with performance comparison on the MultiWOZ dataset.
\end{itemize}

\section{Discussion on existing DST metrics}
\label{sec:evluation_metric}
% In this section, we describe joint and slot accuracy along with their limitations.
\subsection{Joint goal accuracy}
\label{sec:joint_acc}
\begin{figure}[t]
    \begin{center}
        \includegraphics[scale=0.58,trim=55 128 160 42, clip]{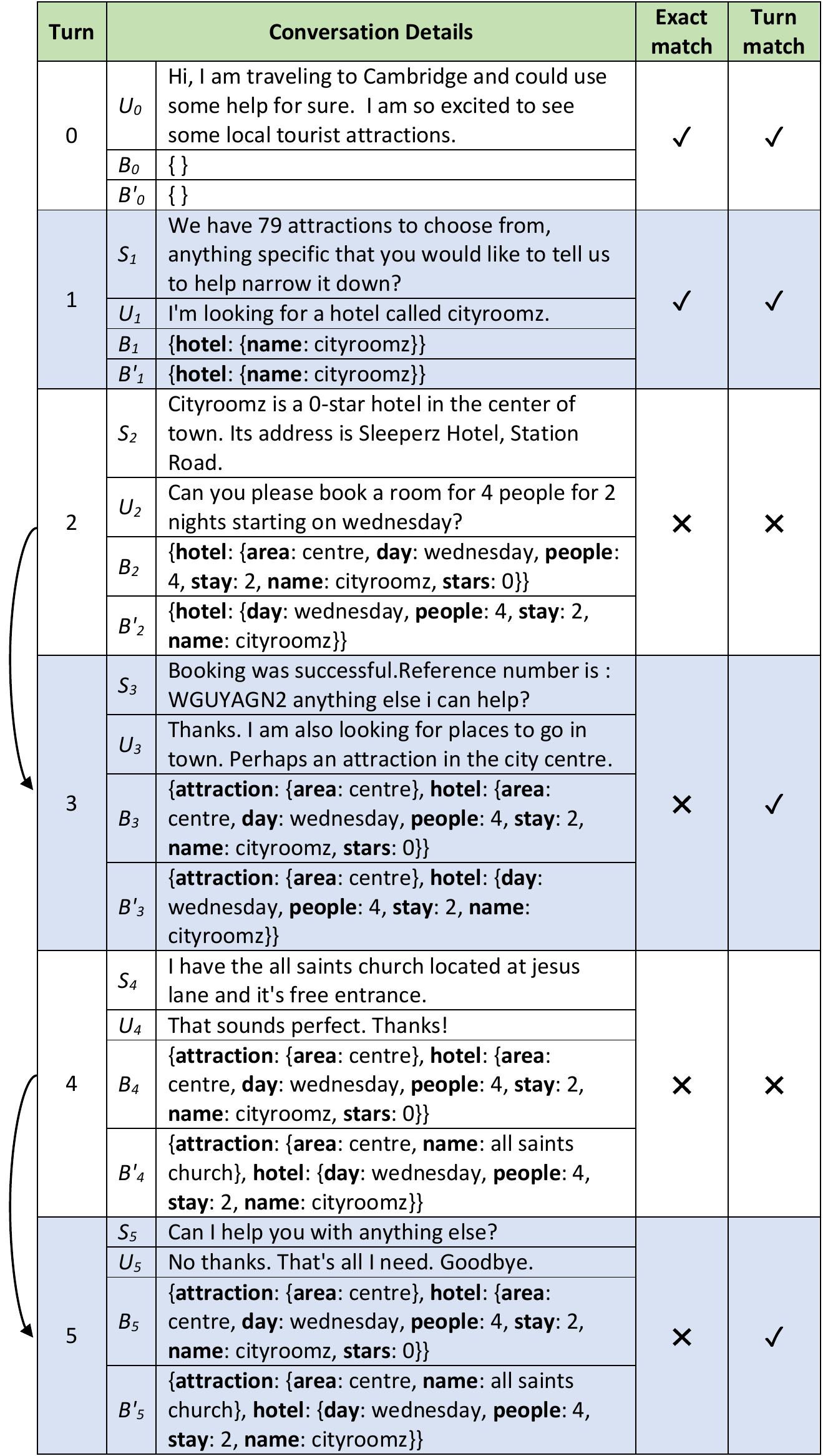}
    \caption{\small{Illustration of DST task. ``Exact Match'' compares Ground truth belief state $B_t$ and Predicted belief state  $B'_t$. ``Turn Match'' indicates the correctness of turn-level non-cumulative belief state prediction. Arrows represent the propagation of errors.}}
    % \caption{\small{Illustration of DST task. `Exact Match'' compares ground truth $B_t$ and predicted $B'_t$. ``Turn Match'' indicates the correctness of turn-specific prediction. Arrows represent the propagation of errors.}}
    %(dialogue-id SNG1070)
    \vspace{-.3in}
    \label{fig:dst-pr}
    \end{center}
\end{figure}

Joint accuracy or joint goal accuracy (JGA) checks whether the set of predicted belief states exactly matches the ground truth for a given user turn \cite{dstc2, trade}.
Let $B_t$ and $B'_t$ be the set of ground-truth and predicted belief states at turn $t$. Then the prediction of turn $t$ is considered to be correct if and only if $B_t$ exactly matches $B'_t$. Fig. \ref{fig:dst-pr} shows an illustration of the predicted belief state where the predictions of $B'_t$ are generated using SOM-DST \cite{som-dst}. In the example, there are 2 out of 6 correct predictions of $B'_t$ that result in a JGA score of 33.33\% for the whole conversation.

Although joint goal accuracy is a convenient metric to evaluate DST, it has certain limitations. The main source of the issue is the cumulative nature of ground-truth $B_t$. As a result, once a misprediction has occurred, it is difficult to get back a correct prediction in subsequent turns. For example, in Fig. \ref{fig:dst-pr}, the prediction goes wrong in Turn $2$ which affects all the later predictions. So, it is very likely to get a JGA of zero if the model somehow mispredicts the first turn. Therefore, JGA can undermine the true potential of a DST model and provide an underestimated performance.

In addition, JGA does not take into account turn-level performances. For instance, in Fig. \ref{fig:dst-pr}, Turn $3$ and $5$ are locally correct but JGA will mark them 0 since $B_t$ and $B'_t$ has not matched exactly. Normally, it is expected that increasing the exact matches will also reflect in turn-level matches. But we observed that sometimes increasing exact matches can decrease turn-level matches mainly due to annotation inconsistencies. So, one should be careful while using only joint accuracy for model selection. Besides, the available DST datasets (like MultiWOZ) contain a lot of annotation errors \citep{multiwoz-2.2}. For example in turn $4$, the model has predicted the intent \textit{(attraction, name, all saints church)}. Although the prediction looks rational, the triplet is absent in the ground-truth. So, if a mismatch occurs due to an annotation error, it is highly probable that all the subsequent turns will be marked incorrect leading to an underestimated performance.

Hence, using joint goal accuracy for evaluating DST works fine if there are no annotation errors and the sole purpose is to improve the prediction of cumulative belief state. Otherwise, there is a need to include turn-level performance in order to obtain a fair evaluation of a DST model.

\subsection{Slot Accuracy}
\label{sec:slot_acc}
Slot accuracy (SA) is a relaxed version of JGA that compares each predicted \textit{(domain, slot, slot-value)} triplet to its ground-truth label individually \cite{trade}. Let $S$ be the set of unique domain-slot pairs in the dataset. Let $B_t$ and $B'_t$ be the set of ground-truth and predicted belief states respectively. Then slot accuracy at turn $t$ is defined as
\begin{equation}
    \label{eqn1}
    SA = \frac{|S| - |X| - |Y| + |P \cap Q|}{|S|},
\end{equation}
where $X=(B_t \setminus B'_t)$, $Y=(B'_t \setminus B_t)$, $P$ is the set of unique domain-slot pairs from $X$, and $Q$ is the set of unique domain-slot pairs from $Y$. Basically, in Equation \ref{eqn1},  $|X|$ and $|Y|$ represent the number of false negatives and false positives respectively. Note that if the value of a ground-truth domain-slot pair is wrongly predicted then this misprediction will be counted twice (once in both $X$ and $Y$). The term $|P \cap Q|$ in the above equation helps to rectify this overcounting. In MultiWOZ, the value of $|S|$ is 30. For Turn 2 in our running example, since $|B_1 \setminus B'_1|=2$ and $|B'_1 \setminus B_1|=0$, slot accuracy is equal to $\frac{(30-2-0-0)}{30}$ i.e. 93.33\%. Slot accuracy for the entire conversation in Fig. \ref{fig:dst-pr} is 94.44\%.

The value of slot accuracy can be very misleading. For instance, even if the prediction of Turn 2 is wrong in Fig. \ref{fig:dst-pr}, we get a slot accuracy of 93.33\% which is extremely high. Basically, slot accuracy overestimates the DST performance. Let us exhibit this fact by considering the case where we predict nothing for all turns i.e. $B'_t=\emptyset,  \forall t$. Then, slot accuracy simplifies to $\frac{|S| - |B_t|}{|S|}$. It is natural that $|B_t|<<|S|$ because a conversation will typically have only a small number of domain-slot pairs \textit{live} at any time. As a result, slot accuracy remains on the higher side ($\approx$ 81\% for MultiWOZ 2.1) even if we predict nothing. For datasets with a larger number of domain/slots, since $|S|$ is large, slot accuracy will be close to 1 for almost all scenarios. Thus, slot accuracy is a poor metric to evaluate DST.

% With the theme of conversational AI gaining more popularity, we anticipate larger datasets with many more domains and slots to come up in near future. Such datasets will have larger domain-slot pairs and hence the issue with slot accuracy highlighted above will become more prominent.

\subsection{Average Goal accuracy}
\label{sec:avg_acc}
Average goal accuracy (AGA) is a relatively newer metric proposed to evaluate the SGD dataset \cite{sgd}. Here, the slots that have a non-empty assignment in the ground-truth dialogue state are only considered during evaluation. Let $N_t \subseteq B_t$ be the set of ground-truth triplets having non-empty slot-values. Then AGA is computed as $\frac{|N_t \cap B'_t|}{|N_t|}$ where $B'_t$ is the predicted belief state for turn $t$. The turns having $N_t=\emptyset$ are ignored during the computation of AGA. In Fig. \ref{fig:dst-pr}, AGA for turn 2 is 4/6, and 76.19\% for the entire conversation. 

This metric has mainly two limitations. Firstly, AGA is only recall-oriented and thereby does not consider the false positives. Ignoring the false positives makes this metric insensitive to extraneous triplets in the predicted belief state. However, this issue can be easily addressed by redefining AGA as $\frac{|N_t \cap B'_t|}{|N_t \cup B'_t|}$. But there still exists a second major problem with AGA. Note that even if a turn is completely wrong, AGA for that turn can still be higher because of the correct predictions in the previous turns. For example, even if turn 2 and 4 are incorrect, we get an AGA of 4/6 and 5/7 respectively which clearly indicates an overestimation.

\section{Flexible Goal Accuracy}
From the previous discussion, it is evident that despite a few limitations, joint goal accuracy is superior to the other two metrics. This is why with the objective to obtain a better evaluation metric for DST, we address the shortcomings of JGA by proposing a new metric called Flexible goal accuracy (FGA). The description of FGA is presented in the next part of this section, whereas its working is described as a pseudo-code in Algo. \ref{algo:fga}.

\begin{algorithm}[t]
\DontPrintSemicolon
\SetAlgoLined 
%\SetAlgoNoLine
\KwIn{$B = $ list of groun-truth belief states, $B'$ = list of predicted belief states, $N$ = \#turns}
\KwOut{Flexible goal accuracy}

$T = \{0, 1, \ldots, N-1\}$ , $t_{err} \gets -\infty$, f = 0\; 
\For{$t \in T$}{
    $w \gets 1$\;
    \If{$B_t \neq B'_t$}{
        \If{$t = 0$}{
            \tcc{Type 1 error}
            $w \gets 0$ , $t_{err} \gets t$  \;
        }
        \Else{
            $T_t \gets B_t \setminus B_{t-1}$\; 
            $T'_t \gets B'_t \setminus B'_{t-1}$\; 
            \If{$T'_t \not\subseteq B_t$ or $T_t \not\subseteq B'_t$}{
                \tcc{Type 1 error}
                $w \gets 0$, $t_{err} \gets t$\; 
            }
            \Else{
                \tcc{Type 2 error}
                $x \gets (t - t_{err})$\; 
                $w \gets 1-\exp(-\lambda x)$\; 
            }
        }
    }
    $f \gets f + w$\;
}
\KwRet $f/N$\;
 \caption{FGA for single conversation}
 \label{algo:fga}
 %\vspace{-.05in}
\end{algorithm}

\begin{table*}[ht]
\begin{small}
\centering
\begin{tabular}{lrrrrrrrrrr}
% \hline \textbf{Model} & \textbf{\#Turns} & \textbf{\#M1} & \textbf{\#M2} & \textbf{JGA} & \textbf{SA} & \textbf{AGA} & \textbf{FGA@0.25} & \textbf{FGA@0.5} & \textbf{FGA@0.75} & \textbf{FGA@1} \\ \hline
\hline \textbf{Model} & \textbf{\#Turns} & \textbf{\#M1} & \textbf{\#M2} & \textbf{JGA} & \textbf{SA} & \textbf{AGA} & \textbf{FGA\textsubscript{0.25}} & \textbf{FGA\textsubscript{0.5}} & \textbf{FGA\textsubscript{0.75}} & \textbf{FGA\textsubscript{1}} \\ \hline
TRADE & 7368 & 3600 & 5287 & 48.86\% & 96.96\% & 88.79\% &  56.58\% & 61.19\% & 64.16\% & 66.18\% \\
Hi-DST  & 7368 & 3622 & 5903 & 49.16\% & 96.70\% & 90.74\% & 61.31\% & 67.69\% & 71.47\% & 73.91\% \\
SOM-DST & 7368 & 3912 & 6084 & 53.09\% & 97.36\% & 91.71\% & 64.94\% & 71.04\% & 74.61\% & 76.88\% \\
Trippy & 7368 & 3926 & 5875 & 53.28\% & 97.30\% & 90.75\% & 63.24\% & 68.67\% & 71.97\% & 74.13\% \\
\hline
\end{tabular}
\caption{\label{tbl:result} Comparison of DST metrics. ``M1'' and ``M2'' represents exact and turn-level matches respectively. ``FGA\textsubscript{x}'' indicates the FGA value calcualated using $\lambda$=x.
}
\label{table:result}
\end{small}
\vspace{-.1in}
\end{table*}

For a given a turn $t$, an error in belief state prediction (i.e. $B_t \neq B'_t$) can occur in two ways: 1) the source of the error is turn $t$ itself i.e. the turn-level prediction is wrong, 2) the turn-level prediction of turn $t$ is correct but the source of the error is some earlier turn $t_{err} \prec t$. FGA works differently from JGA only for type 2 errors. Unlike JGA, FGA does not penalize type 2 errors completely. It assigns a penalized score based on the distance between the error turn ($t_{err}$) and the current turn ($t$) and the penalty is inversely proportional to this distance ($t-t_{err}$). The main idea is to forget the mistakes with time in order to attain a fair judgment of a DST model offline. 

We decide the correctness of a turn-level match using the logic shown in line 10 of Algo. \ref{algo:fga}. A turn $t>0$ is locally correct if ($T'_t \subseteq B_t$ and $T_t \subseteq B'_t$) where $T_t=B_t\setminus B_{t-1}$ and $T'_t = B'_t \setminus B'_{t-1}$. In other words, a turn-level or local match indicates that all the intents shown by the user in a particular turn have been correctly detected without any false positives. Just comparing $T_t$ and $T'_t$ to check a turn-level or local match can be erroneous because it will not credit the model for error corrections. For the penalty function, we use the CDF of exponential distribution (shown in Line 14 of Algo. \ref{algo:fga}) parameterized by $\lambda$ where $\lambda \geq 0$. Clearly, the strictness of FGA is inversely proportional to $\lambda$. Note that $\lambda = 0$ will reduce FGA to JGA (strict metric) whereas $\lambda \to \infty$ will report only the accuracy on turn-level matches (relaxed metric). Finding the appropriate $\lambda$ for a specific DST task should be done carefully in order to match the desired evaluation criteria. However, we can take a theoretical stand and approximate the hyper-parameter value as $\lambda=-ln(1-p)/t_{f}$ where $t_{f}$ is the number of turns that it will take to forget a mistake by factor $p$ where $(0\leq p<1)$. For example, if $t_{f}$=6 and $p$=0.95, then $\lambda$=0.499. So, the strictness of FGA is directly proportional to $t_f$ and inversely proportional to $p$. If the dataset is clean, one can alternatively find the best $\lambda$ through a human evaluation, although it would require additional human effort. Hence, we can flexibly set the strictness criteria of FGA through the hyper-parameter $\lambda$ according to our requirement.

In our running example (Fig. \ref{fig:dst-pr}), the FGA score for each turn with $\lambda=0.5$ is \{1, 1, 0, 0.39, 0, 0.39\} which results in a FGA score of 46.33\% for the entire conversation. We can observe two things from these numbers. Firstly, it is not overestimating in comparison to SA and AGA. Secondly, it gives a better estimate than JGA in keeping track of both exact and turn-level matches simultaneously. Hence, FGA can provide a relatively balanced estimate than the existing metrics even in the presence of annotation errors and inconsistencies.

% \comg{The level of strictness while evaluating DST may vary depending on the task or dataset. FGA gives the flexibility to set the strictness criteria through the hyper-parameter $\lambda$. As discussed earlier, setting $\lambda=0$ reduces FGA to JGA (strict metric) while setting $\lambda=\infty$ or some large value reduces FGA to turn-level accuracy (relaxed metric). Finding the appropriate $\lambda$ for a specific DST task should be done carefully in order to match the desired evaluation criteria. However, we can take a theoretical stand and approximate the hyperparameter value as $\lambda=-ln(1-p)/t$ where $t$ is the number of turns that it will take to forget a mistake by factor $p$ where $(0\leq p<1)$. For example, if $p$=0.95 and $t$=6, then $\lambda$=0.499. So, higher value of $t$ will result in a stricter metric while lowering the value will relax it. If the dataset is clean, one can alternatively find the best $\lambda$ through a human evaluation, although it would require additional human effort.}

\section{Result and Analysis}
In this section, we report the performance of FGA along with the other metrics on four different DST models: TRADE \citep{trade}, Hi-DST \citep{hi-dst}, SOM-DST \citep{som-dst}, and Trippy \citep{trippy}. We use the MultiWOZ 2.1 dataset \citep{multiwoz-2.1} as most of the recent progress in DST are showcased on this dataset. The results are reported in Table \ref{table:result}. Since the MultiWOZ dataset covers many domains (hotel, restaurant, taxi, train, attraction) where each domain may have different levels of tolerance (intuitively train, taxi booking may be strict whereas information seeking about attraction, restaurant domains may be lenient), an overall common/single strictness setting for the entire dataset may be difficult to reach at. Hence, we reported the FGA score for multiple values of hyper-parameter $\lambda$ rather than showing the result for a single value. For the same reason, we did not try to find the best $\lambda$ for evaluating the MultiWOZ dataset.

% From Table \ref{table:result}, we can observe that Trippy has the best JGA. Currently, most of the state-of-the-art DST performances are shown using Trippy. However, we can notice that Trippy does not have the same performance gain for turn-level matches. This behavior can be a side-effect of boosting the JGA using intricate featurization. In contrast, Hi-DST optimizes explicitly for turn-level non-cumulative belief states, thereby achieving better turn-level accuracy than JGA.  Among the four models, SOM-DST performs well for both objectives because of their sophisticated selective overwrite mechanism. Now, by comparing the numbers of Table \ref{table:result}, we can infer that FGA does a better job in providing a fair estimate while considering both exact and turn-level matches. Moreover, we can also notice that FGA acts as a better discriminator of DST models in comparison to the existing metrics.
From Table \ref{table:result}, we can observe that Trippy has the best JGA. Currently, most of the state-of-the-art DST performances are shown using Trippy. However, we can notice that Trippy does not have the same performance gain for turn-level matches. It has lesser turn-level matches than SOM-DST and Hi-DST. This behavior of Trippy can be a side-effect of boosting the JGA using its intricate featurization. In contrast, Hi-DST optimizes explicitly for turn-level non-cumulative belief states, thereby achieving better turn-level accuracy at the expense of JGA.  Among the four models, SOM-DST performs well for both objectives because of their sophisticated selective overwrite mechanism. Now, by comparing the numbers of Table \ref{table:result}, we can infer that FGA does a better job in providing a fair estimate while considering both exact and turn-level matches. Moreover, we can also notice that FGA acts as a better discriminator of DST models in comparison to the existing metrics.

% \textbf{Human Evaluation:} We conducted a human evaluation on 100 randomly picked conversations from the MultiWOZ 2.1 test data. For each turn in a conversation, we provided the system and user utterance along with the ground-truth and predicted belief state. The predictions were generated using SOM-DST. Evaluators were asked to provide a single rating (on a scale of 1-5) for an entire conversation based on the quality of the predicted belief states in keeping track of user intent. Pearson correlation coefficient of JGA and FGA with human ratings came out to be 0.62 and 0.63 respectively. Whereas we got 0.67 and 0.69 for SA and AGA respectively. We believe that the misprediction penalties used by the humans gave SA and AGA an advantage. Moreover, in many instances the evaluators found the predictions to be more appropriate than the ground-truth. As a consequence, it gave AGA further advantage since it is recall-oriented.

\textbf{Human Evaluation:} We conducted a human evaluation involving 11 evaluators on 100 randomly picked conversations from the MultiWOZ 2.1 test data. For each turn in a conversation, we provided the system and user utterances along with the ground-truth and predicted belief states. The predictions were generated using SOM-DST. For each conversation, the evaluators were asked to report their satisfaction ($1$) or dissatisfaction ($0$) with the performance of the model in keeping track of user intent throughout the conversation. Pearson correlation coefficient of JGA and FGA (with $\lambda = 0.5$) with human ratings came out to be 0.33 and 0.37 respectively. This shows that FGA is slightly better correlated than JGA with human evaluation.

\section{Conclusion}
% In this work, we analyzed the limitations of existing DST metrics. We argued that joint accuracy can underestimate the power of a DST algorithm, whereas slot and average goal accuracy can overestimate it. We addressed the issues of joint accuracy by introducing Flexible goal accuracy (FGA) which tries to give partial credit to mispredictions that are locally correct. We justified that FGA provides a relatively balanced estimation of DST performance with the flexibility to customize the evaluation goal. 
% % We also showed that FGA has a slightly better correlation with human ratings than JGA. 
% In conclusion, FGA is a practical and insightful metric that can be useful to evaluate future DST models.
In this work, we analyzed the limitations of existing DST metrics. We argued that joint accuracy can underestimate the power of a DST algorithm, whereas slot and average goal accuracy can overestimate it. We addressed the issues of joint accuracy by introducing Flexible goal accuracy (FGA) which tries to give partial credit to mispredictions that are locally correct. We justified that FGA provides a relatively balanced estimation of DST performance along with better discrimination property. In conclusion, FGA is a practical and insightful metric that can be useful to evaluate future DST models.

\bibliography{anthology,mybib}
\bibliographystyle{acl_natbib}

\appendix

\section{Appendix}
\label{sec:appendix}
\subsection{MultiWOZ Dataset}
MultiWOZ \citep{multiwoz} is a popular DST corpus that contains both single and multi-domain conversations. For this work, we used MultiWOZ 2.1 \citep{multiwoz-2.1} which is an updated version of the original MultiWOZ 2.0 dataset. In addition to the original dataset, MultiWOZ 2.1 contains fixes to some noisy annotations. Table \ref{tbl:app1} shows few elementary statistics of the dataset.
\begin{table}[ht!]
\begin{small}
\centering
\begin{tabular}{lrrr}
\hline \textbf{Data} & \textbf{\#Conversations} & \textbf{\#Turns} & \textbf{Avg. turns}  \\ \hline
Train & 8420 & 56668 & 6.73 \\
Dev & 1000 & 7374 & 7.37 \\
Test & 999 & 7368 & 7.37 \\
\hline
\end{tabular}
\caption{\label{tbl:app1} Elementary statistics of MultiWOZ 2.1 dataset. ``Avg. turns'' indicate average turns per conversation.
}
\end{small}
\end{table}

\subsection{Result generation procedure}
We generated results for four DST models - Trade \citep{trade} \footnote[2]{
\href{https://github.com/jasonwu0731/trade-dst}{github.com/jasonwu0731/trade-dst}}, Hi-DST \citep{hi-dst} \footnote[3]{
\href{https://github.com/SuvodipDey/Hi-DST}{github.com/SuvodipDey/Hi-DST}}, SOM-DST \citep{som-dst} \footnote[4]{
\href{https://github.com/clovaai/som-dst}{github.com/clovaai/som-dst}}, and Trippy \citep{trippy} \footnote[5]{
\href{https://gitlab.cs.uni-duesseldorf.de/general/dsml/trippy-public}{gitlab.cs.uni-duesseldorf.de/general/dsml/trippy-public}}. We used their official code to train them on MutiWOZ 2.1 dataset. All four models generate an inference file that contains the predicted belief states for the test set. We used these inference files to compute the values of different metrics shown in Table \ref{table:result}. As we trained all the models from scratch, the results may not be exactly the same as those reported in the original paper.

\subsection{Human evaluation format}

\begin{figure*}[ht!]
    \begin{center}
        \includegraphics[scale=0.9,trim=63 100 68 65, clip]{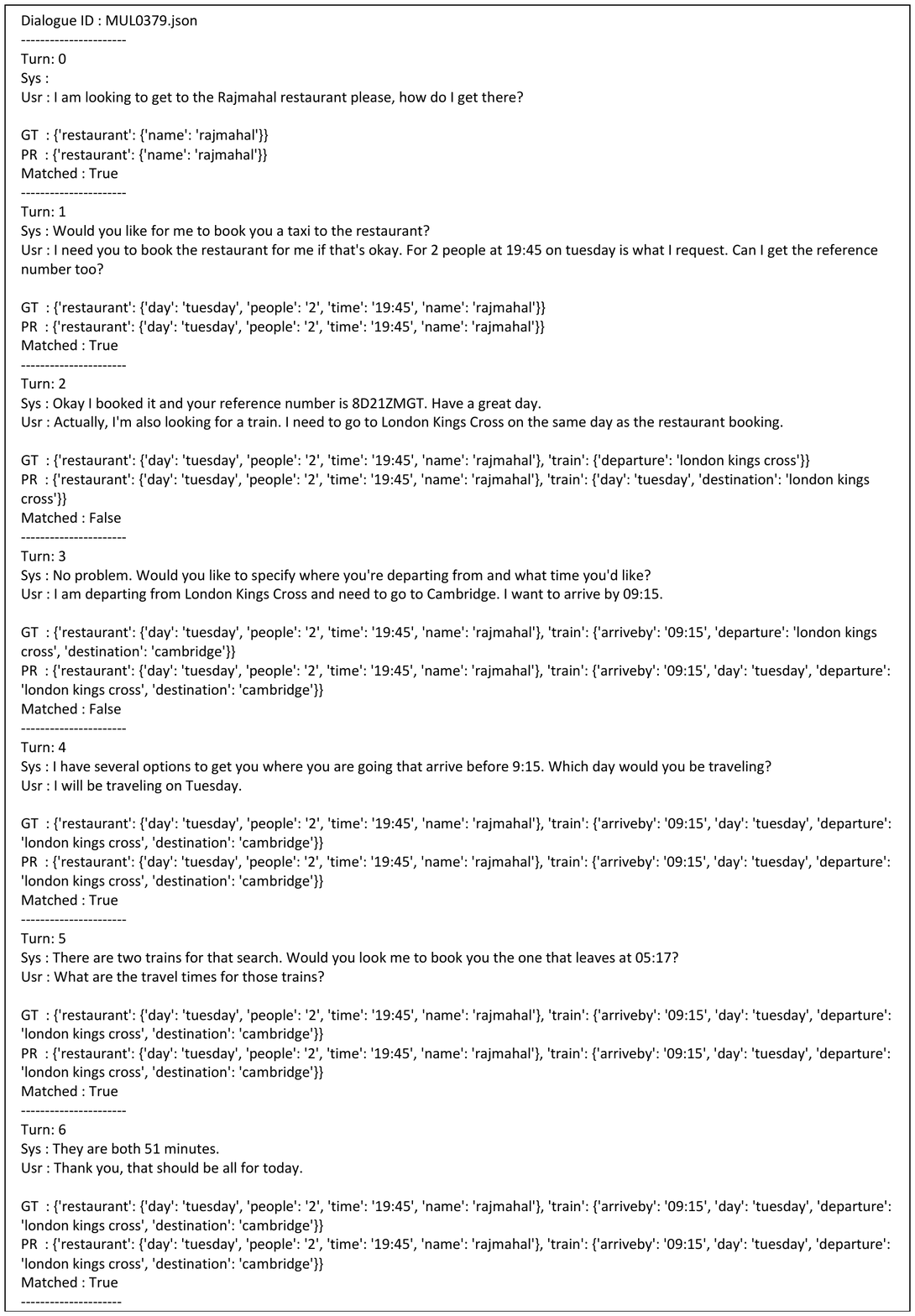}
    \caption{Data format for human evaluation}
    % \caption{\small{Illustration of DST task. `Exact Match'' compares ground truth $B_t$ and predicted $B'_t$. ``Turn Match'' indicates the correctness of turn-specific prediction. Arrows represent the propagation of errors.}}
    %(dialogue-id SNG1070)
    \vspace{-.1in}
    \label{fig:sample-eval}
    \end{center}
\end{figure*}
For each randomly picked conversation for human evaluation, we prepared a file that logged the utterances, ground-truth, and predicted belief state for each turn. Additionally, we indicated whether the ground truth exactly matched the predicted belief state to speed up the evaluation process. A sample file format is shown in Fig. \ref{fig:sample-eval}. 
\end{document}